\documentclass[sigconf, authorversion, nonacm]{acmart}
\usepackage{hyperref}       
\usepackage{url}            
\usepackage{booktabs}       
\usepackage{amsfonts}       
\usepackage{nicefrac}       
\usepackage{microtype}      
\usepackage{amsmath}
\usepackage{natbib}
\usepackage{multicol}
\usepackage{multirow}
\usepackage{graphicx}
\usepackage{enumitem}
\usepackage{wrapfig}
\usepackage{tabularx,ragged2e}
\usepackage{longtable}
\usepackage{adjustbox}
\usepackage[algoruled,boxed,lined]{algorithm2e} 
\usepackage{comment}
\usepackage{caption}
\usepackage{tabularx,ragged2e}
\usepackage{longtable}
\usepackage{soul}
\usepackage[algoruled,boxed,lined]{algorithm2e} 
\usepackage{comment}
\usepackage{caption}

\newcolumntype{L}{>{\RaggedRight\arraybackslash}X} 
\newcolumntype{T}{>{\RaggedRight\arraybackslash}X}

\AtBeginDocument{%
  \providecommand\BibTeX{{%
    \normalfont B\kern-0.5em{\scshape i\kern-0.25em b}\kern-0.8em\TeX}}}

\begin{document}

\title{Multi-Intent Detection in User Provided Annotations for Programming by Examples Systems}

\author{Nischal Ashok Kumar}
\authornote{Both authors contributed equally to the paper}
\authornote{Work done during internship at IBM Research}
\affiliation{%
 \institution{UMass Amherst}
 \country{United States}
}
 \email{nashokkumar@umass.edu}

\author{Nitin Gupta}
\authornotemark[1]
\affiliation{%
 \institution{IBM Research}
  \country{India}
}
 \email{ngupta47@in.ibm.com}

\author{Shanmukha Guttula}
\affiliation{%
 \institution{IBM Research}
  \country{India}
}
 \email{shagutt1@in.ibm.com}

\author{Hima Patel}
\affiliation{%
 \institution{IBM Research}
  \country{India}
}
 \email{himapatel@in.ibm.com}


\settopmatter{printacmref=false}


\begin{abstract}
In mapping enterprise applications, data mapping remains a fundamental part of integration development, but its time consuming. An increasing number of applications lack naming standards, and nested field structures further add complexity for the integration developers. Once the mapping is done, data transformation is the next challenge for the users since each application expects data to be in a certain format. Also, while building integration flow, developers need to understand the format of the source and target data field and come up with transformation program that can change data from source to target format. The problem of automatic generation of a transformation program through program synthesis paradigm from some specifications has been studied since the early days of Artificial Intelligence (AI). Programming by Example (PBE) is one such kind of technique that targets automatic inferencing of a computer program to accomplish a format or string conversion task from user-provided input and output samples. To learn the correct intent, a diverse set of samples from the user is required. However, there is a possibility that the user fails to provide a diverse set of samples. This can lead to multiple intents or ambiguity in the input and output samples. Hence, PBE systems can get confused in generating the correct intent program.  In this paper, we propose a deep neural network based ambiguity prediction model, which analyzes the input-output strings and maps them to a different set of properties responsible for multiple intent. Users can analyze these properties and accordingly can provide new samples or modify existing samples which can help in building a better PBE system for mapping enterprise applications. 


\end{abstract}




\keywords{Programming by Example, Program Synthesis, Ambiguity, Multiple Intent}



\maketitle

\section{Introduction}
String Transformation in mapping enterprise applications refers to the specific paradigm in the domain of Programming by Example (PBE) approaches, where a computer program learns to capture user intent, expressed through a set of input-output pairs, from a pre-defined set of specifications and constraints \cite{manna1975knowledge,devlin2017robustfill,gulwani2011automating,polozov2015flashmeta,shu2017neural}. The set of specifications and constraints is expressed through the Domain-Specific Language (DSL) which consists of a finite number of atomic functions or string expressions that can be used to formally represent a program for the user to interpret. Most of the PBE systems \cite{gulwani2011automating,polozov2015flashmeta,menon2013machine,alur2013syntax} for string transformation use ranking mechanisms that are either built using heuristics or learned using historical data. These kinds of ranking systems are designed to capture the following two important characteristics: small length and simpler programs. Such kind of ranking system mostly depends on the quality and number of input and output (I/O) annotation samples to learn better program. The quality of I/O samples denotes how good the I/O samples are to generate a single intent output.

The number of given I/O annotation samples can vary depending on the user intent, but the fewer the better for the user (as the user has to provide less annotations). Therein lies the challenge of learning correct intent i.e., if examples are too few, then many possible DSL functions can satisfy them, and picking one intent (or program) arbitrarily or based on some ranking mechanism that satisfies simplicity and smaller length criteria, might lead to non-desired intent program. This might yield a solution that works well only on the given I/O samples but not on unseen samples. Similarly, the quality of I/O samples (irrespective of high I/O samples count) plays an important role in generating the correct intent program. The above two challenges are critical for PBE kind of systems to understand the user's intent by analyzing the given I/O samples. This can lead to a sub-optimal program that works on seen data but does not give desired outputs for unseen data. Hence, it is important to understand whether the given I/O samples capture the user's desired intent correctly or not. 
\begin{table}[t] \scriptsize
\centering
 \resizebox{.49\textwidth}{!}{
\begin{tabular}{|c|c|c|c|c|}
\hline
     \multicolumn{2}{|c}{\textbf{Train}}& \multicolumn{3}{|c|}{\textbf{Test}}\\\hline
    \textbf{Input} & \textbf{Output} & \textbf{Input}& \textbf{Generated Output}&\textbf{ GT Output} \\\hline
    ABCD\_12 & 12 & & &\\
    BDJ\_535 & 535 & B\_DS2345 & 2345 & DS2345 \\
    GE\_443 & 443 & & &\\
    \hline
    07-07-1999 & 07 & & & \\
    02-02-1955 & 02 & 09-07-1995 & 09 & 07 \\
    10-10-2002 & 10 & & & \\ 
    \hline
  
    \end{tabular}
    }
    \caption{Examples demonstrating the challenges where I/O annotations can lead to multiple intents.}
        \label{tb:challenges}   
\vspace{-4mm}
\end{table}
\begin{table}[t] \scriptsize
\centering
\begin{tabularx}{\linewidth}{|l|l|L|}
\hline
     \multicolumn{2}{|c|}{\textbf{Train}}& \multirow{2}{*}{\textbf{Program(s)}}\\\cline{1-2}
    \textbf{Input} & \textbf{Output}&\\\hline
     & & Prog1 -  Extract number substring after ``\_"\\
    ABCD\_12 & 12 & Prog2 -  Extract string after ``\_"\\
    BDJ\_535 & 535 & Prog3 -  Extract substring between regex-[A-Z]+[\_] and last indices\\
    GE\_443 & 443 & Prog4 -  Extract substring between first regex-[0-9] and last indices\\
    && ProgN -  .....\\
    \hline
   & & \st{Prog1 -  Extract number substring after ``\_"}\\
    ABCD\_12 & 12 & Prog2 -  Extract string after ``\_"\\
    BDJ\_535 & 535 & Prog3 -  Extract substring between regex-[A-Z]+[\_] and last indices\\
    GE\_D443 & D443 & \st{Prog4 -  Extract substring between first regex [0-9] and last indices}\\
    && ProgN -  .....\\
    \hline
    \end{tabularx}
    \caption{Examples demonstrating the I/O pairs that can lead to different possible programs which generate multiple outputs. The strike lines show the programs that have been removed after modifying sample 3 in example 2.}
        \label{tb:challenges2}  
\vspace{-8mm}
\end{table}
For illustration, let's take an example shown in Table \ref{tb:challenges}. "Train" columns denote the columns representing I/O samples used to generate a transformation program and "Test" columns denote the input sample column which is passed to the transformation program to generate an output. GT output column denotes the actual desired output. For each example, the user provides 3 I/O samples to generate a transformation program using \cite{polozov2015flashmeta}. In the first example, user intent is to extract substring after ''\_"  character, but here PROSE system learns the program which transforms test input ``B\_DS2345" into test output ``2345" (see generated output column), which implies that the system learns to extract last numeric substring, which is different from a user-desired intent. This happens because there can be many possible programs to transform one set of inputs into outputs. Sometimes those programs converge to the same intent and other times it can lead to multiple intents. For example, in Table \ref{tb:challenges2}, for the first set of I/O samples, multiple programs can be possible. For test input sample ``B\_D2S345", where desired output value is ``D2345". However, programs in Program(s) column generate different values for this example, first program generates - ``345", second program -``D2S345", third program - ``D2S345", fourth program - ``345" and so on. This shows that all these consistent programs with I/O samples can lead to multiple intents (or outputs) on unseen data. For the above use case, two clear intents are - (a) Extract numeric substring after ``\_", and second intent is extract substring after ``\_". But if we look at the second row in Table \ref{tb:challenges2}, where we replaced the third sample with a better sample ``GE\_D443 - D443", then automatically first intent program got eliminated from the programs list. Hence, accessing the quality of annotations with respect to single or multiple intents is required for better PBE systems. If the user provides sufficient and single intent specific samples, the system can easily generalize to the rest of the samples. Hence, there is a need for a system that analyses the I/O samples that can help in finding multiple intent issues in annotations. 
This would help in informing the user about multi-intent issues before generating a transformation program.

Therefore, we propose a framework to understand the quality of I/O samples to accurately predict a single confident program. To achieve this goal, we introduce a set of generic properties which helps to find ambiguity/multiple intents in a given set of I/O annotation samples. These properties are generic enough for most of the PBE systems because these properties are designed by analyzing several PBE systems' DSL. We propose a deep learning-based framework to automatically identify the presence of these properties in the annotations. The proposed framework takes a set of I/O samples annotation pairs as input and analyzes those samples together to classify the annotations to these properties. User can utilize this information to enhance the I/O samples, hence, generating more accurate, single intent, simpler and shorter program. In summary, the core contributions of our work are as follows:

\begin{itemize}[leftmargin=*]
\item Multi-Tasking Attention-Based Deep Neural Network to address the issues of input and output annotation quality to generate a program with the correct intent.
\item Defined a set of generic properties after analyzing several PBE systems' DSL that can help to find whether a given set of I/O samples can lead to multiple intents or not. 
\item We present an extensive quantitative analysis of a synthetically generated dataset. We also show the motivation of each module of our proposed framework through an ablation study. 
\item We also demonstrate the impact of detecting multiple intents and correcting them before building any PBE system.
\end{itemize}

\section{Overview of Proposed Methodology}
In this section, we discuss the overview of the proposed methodology, define the set of properties to detect multiple intents, and formally define the problem setting. For any PBE system, I/O samples play an important role in determining the correct intent program. Examples are an ambiguous form of specification: there can be different programs that are consistent with the provided examples, but these programs differ in their behavior on unseen inputs. If the user does not provide a large set of examples or less but good quality samples, the PBE system may synthesize unintended programs, which can lead to non-desired outputs. Hence, there is a need for a framework that can access the quality of I/O samples with respect to multiple intents before generating the program. To access the quality of I/O samples, the most important aspect is to understand how good I/O patterns are for PBE system DSL.

\begin{figure}[ht!]
  \centering
  \includegraphics[scale=0.11]{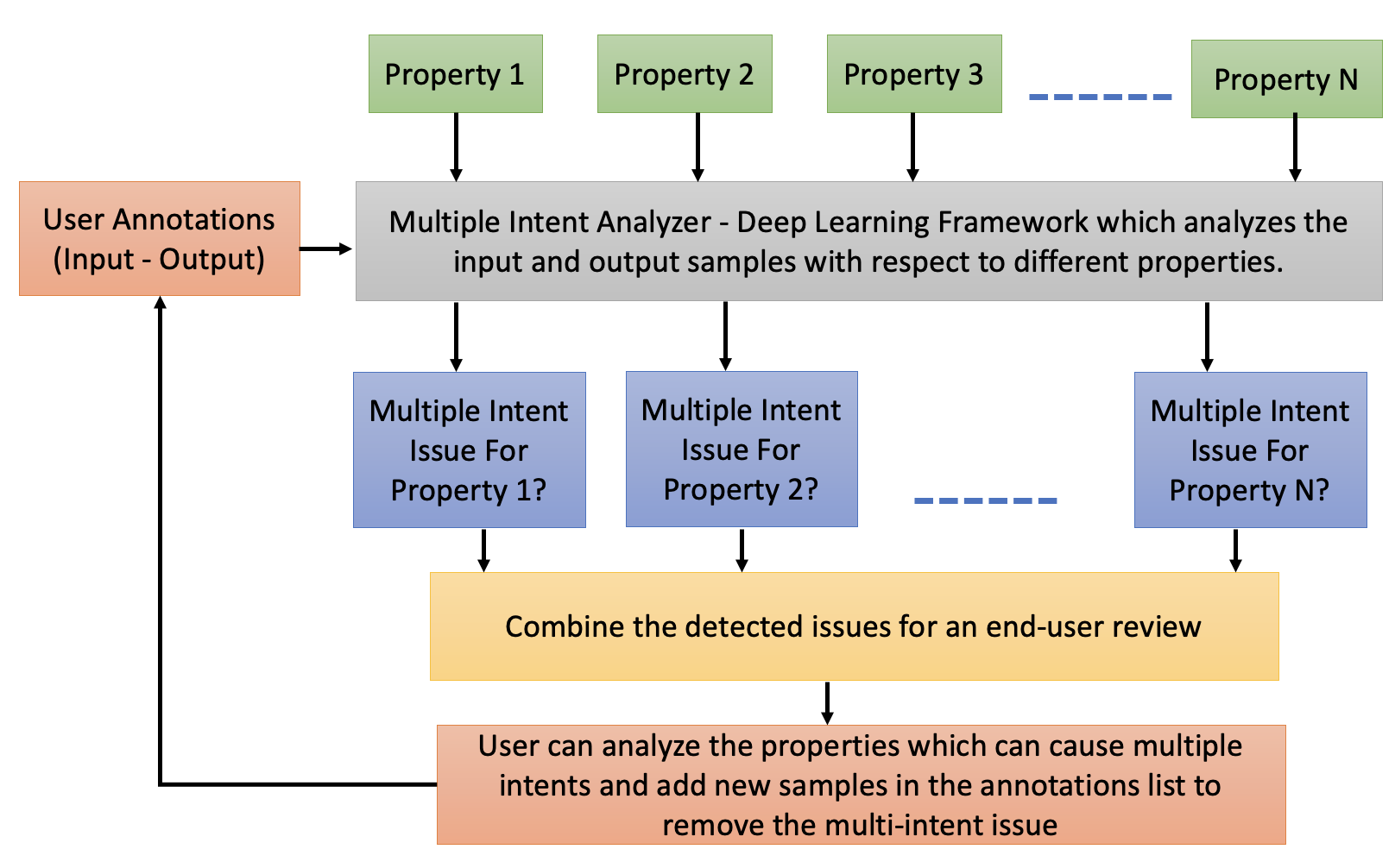}
  \caption{Overview of proposed multi-intents/ambiguity detection framework.}
  \label{fig:framework}
\end{figure}
The proposed framework (Figure \ref{fig:framework}) consists of two major modules, (a) For I/O annotations, defining set of properties which can cause ambiguity or multiple intents - we analyzed several string transformation specific DSL's, and came out with a generic set of properties which helps to identify whether given I/O samples can lead to multiple intents or not. However, the proposed system is generic enough that users can always add a new set of properties based on new functions introduced in DSL which can cause multiple intents, (b) Multiple intent analyzer - we designed a multi-tasking attention-based deep neural network to detect the ambiguity in given I/O samples based on given set of identified properties. The system first analyzes the user I/O annotation samples using the proposed deep learning framework to detect properties that cause multiple intents or ambiguities. In the next step, the user analyzes those detected properties and based on that, add or modifies samples in I/O annotations to improve the overall annotation quality to learn the correct intent program.

In the next section, we will first discuss the properties that will be helpful to decide whether given I/O samples can cause multiple intents or not. In Section 2.2, we will describe the proposed deep learning-based framework which utilizes these properties to find the presence of multiple intents in given annotations.

\begin{table*}[t] \footnotesize
\centering
\begin{tabularx}{\linewidth}{|l|m{2cm}|l|l|m{4.7cm}|L|}
\hline
S.No. & Property Name & Input & Output & Ambiguity& Programs\\ \hline

&\multirow{6}{2.2cm}{Similar Length Ambiguity} & AB\_CD\_123 & CD123 & \multirow{3}{4.7cm}{The desired substring in output ``123" and ``535" from the input are of the same length, hence it is not clear that whether the user wants to extract everything after the second ``\_" or just three characters.} & Concat(SubStr(y1: RelPos('[\_]+', '+0', 0), y2: CPos(-4), case\_type: None), SubStr(y1:CPos(-3), y2:CPos(0)))\\ 
1&&BDG\_SJKL\_535 & SJKL535&&.........\\
&&&&&Concat(Split(sep: \_, index: 1), Split(sep: \_, index: 2))\\
&&&&&\\
\hline
&\multirow{6}{2.2cm}{Exact Position Placement Ambiguity}&Mohan Kumar&Kumar&\multirow{4}{4.7cm}{The desired substring in output ``Kumar" and ``Williams" from the input starts from same position 6, hence it is not clear that whether the user always wants to extract from position 6 or based on some pattern or some other logic.}&SubStr(y1: CPos(6), y2: CPos(0), case\_type: None)\\
2&&&&&........\\
 && Merve Williams & Williams & &SubStr(y1: RelPos('[a-z]+[ ]+', '+0', 0), y2: CPos(0), case\_type: None)\\
 &&&&&\\
 
\hline
&\multirow{6}{2.2cm}{Exact Match Ambiguity}&19-11-1995 & 19/11 &\multirow{3}{4.7cm}{The desired substring in output ``11" and ``11" is the same, hence it is not clear that whether the user always wants to have constant value 11 or wants to extract it from the input string itself.}& Concat(Split(sep: -, index: 0),ConstStr('/11'))\\
3&&&&&.........\\
 && 10-11-2012 & 10/11 & &Concat(Split(sep: -, index: 0),ConstStr('/'),Split(sep: -, index: 1))\\
\hline
&\multirow{6}{2.2cm}{Similar in Token Type Ambiguity}    & A1B-123-A2BD   & 123BD  & \multirow{6}{4.7cm}{The desired substrings in output ``123" and ``53" from the input is of the same type i.e. numeric which can be denoted by regex using  [0-9]+, hence it is not clear that whether the user always wants to extract the numeric value after first ``\_" or any alphanumeric between first ``\_" and second ``\_".}& Concat(SubStr(y1: RelPos('[0-9]+', '-0', 1), y2: RelPos('[0-9]+', '+0', 1), case\_type: None), SubStr(y1: CPos(-1), y2: CPos(0), case\_type: None))\\
4&&1A-53-GGAK     & 53AK &&.........\\
& & &  & &Concat(Split(sep: -, index: 1), SubStr(y1: CPos(-1), y2: CPos(0), case\_type: None))\\
& & &  & &\\
  \hline
&\multirow{6}{2.2cm}{Repeating Characters Ambiguity}   & K 1 TFR 1      & 1   &  \multirow{4}{4.7cm}{The desired substring in output ``1" and ``2" exists at multiple locations location in input, hence it is not clear that whether the user wants to extract this value from position 2 or 8.}& SubStr(y1: CPos(2), y2: CPos(3), case\_type: None) \\
5&&&&&.........\\
&& Y 2 ECN 2 & 2&&.........\\
&  && & &SubStr(y1: CPos(8), y2: CPos(9), case\_type: None)\\

 \hline

\end{tabularx}
\caption{Illustration of different Multi-Intent/Ambiguities detection properties through examples.}
\label{tab:Properties_Example}
\vspace{-8mm}
\end{table*}

\subsection{Properties to Detect Multiple Intent}
\label{prop}
\begin{figure}[t]\scriptsize
  \centering
    \includegraphics[scale =0.08]{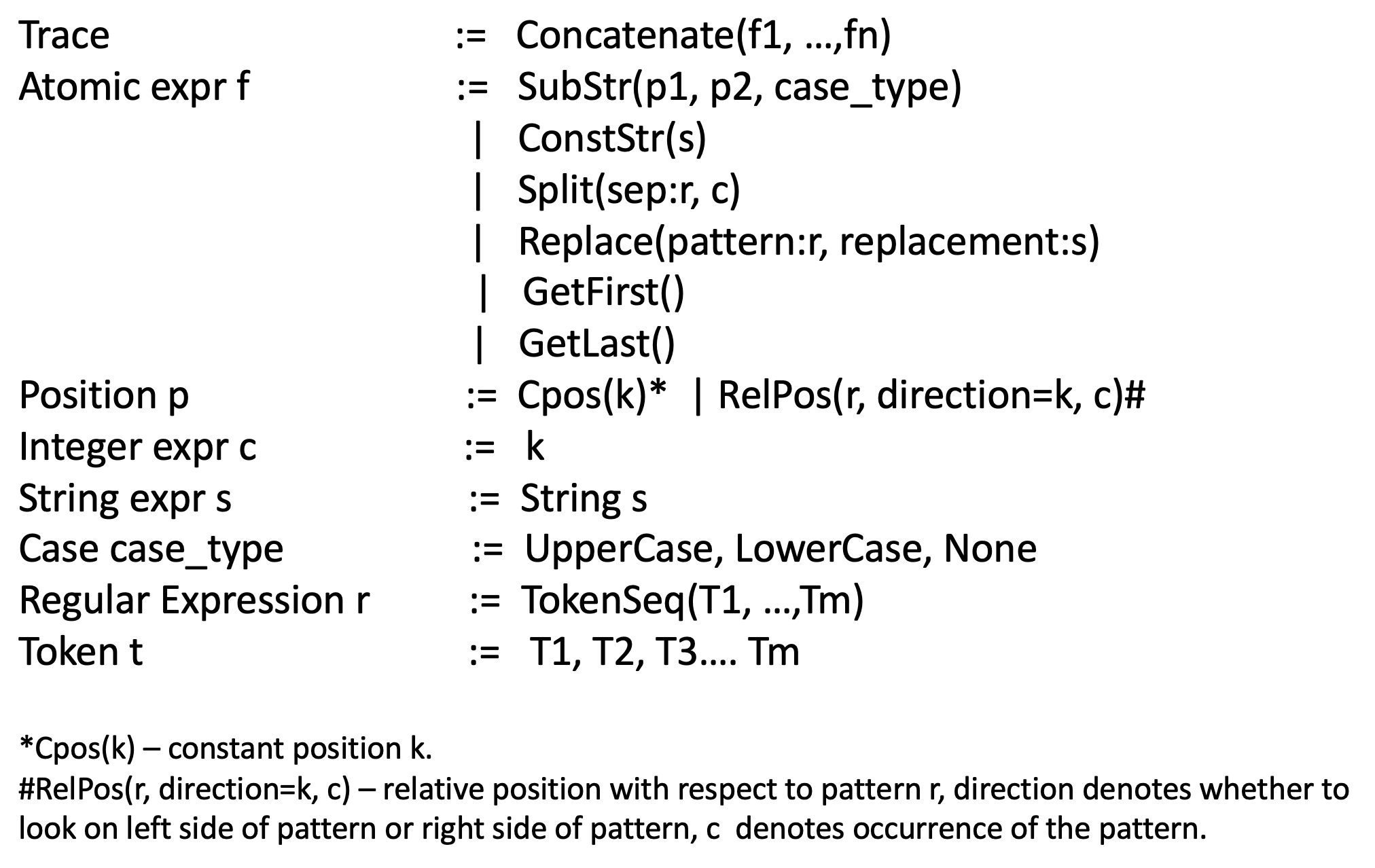}
  \caption{{Domain-Specific Language Semantics.}}
  \label{fig:network_dsl}
  \vspace{-6mm}
\end{figure}

The most important part in finding the ambiguity or possibility of the multiple intents in a given annotation is to analyze the I/O samples for generic characteristics of operators present in the DSL. Mostly, all the DSLs that exist in the literature for string transformation-based PBE systems use similar kinds of operators like split, substring with regex or constant value as an argument, concat, replace, extract first substring, etc. We analyzed several string manipulation-specific DSLs and come out with five generic properties that can help in detecting the multiple intents in the I/O samples. Figure \ref{fig:network_dsl} shows one of the DSL created by combining several other DSL's commonly used operators. There can be other string manipulations operators such as trim, but these are high-level operators and generally doesn't contribute in the   multi-intent scenario. In this paper, we will use the DSL showed in Figure \ref{fig:network_dsl} to illustrate the importance of the defined properties.

Properties of I/O to detect the presence of multiple intents should be tightly bound to the DSL used for the PBE system. At the same time, those properties should also be (1) \textit{concise} enough to capture the implicit or explicit multiple intent and (2) \textit{expressive} enough to allow transformations to be achieved without any confusion in ranking between the programs. Below, we describe the set of 5 properties and the motivation behind their design.

\subsubsection{Similar Length Ambiguity} - This kind of ambiguity can happen when the output substring of all the I/O pairs can be extracted by applying the same DSL operator on corresponding input annotations and have the same length. For example, in Table \ref{tab:Properties_Example}, example 1, following substring or continuous sequence in output ``123" and ``535" are extracted from the similar continuous sequence in input and also are of the same length, hence it is not clear that whether the user wants to extract everything after second ``\_" or just three characters. In terms of DSL, mostly this kind of ambiguity can be possible because of the outputs generated by constant length-based operators like substring with constant positions vs pattern-based operators like split, substring with a pattern. Formally, we can define this property as, given an \emph{example} (\emph{($I_{1}, O_{1}),(I_{2}, O_{2}), ..., (I_{l}, O_{l})$}) satisfies this property when the continues sequence of string in an output matches to the same continuous sequence of string in input and have the same number of characters across that sequence in all the output samples. $I_{l}$ denotes the $l^{th}$ input sample, and $O_{l}$ denotes the corresponding output sample, and $l$  denotes the total I/O samples in one example.

\subsubsection{Exact Position Placement Ambiguity}  - This kind of ambiguity can happen when the output substring of all the I/O pairs can be extracted by applying the same DSL operator on corresponding input annotations and extracted output string always starts or ends on the same position in the input string. For example, in Table \ref{tab:Properties_Example}, example 2,  following substring or continuous sequence in output ``Kumar" and ``Williams" are extracted from the similar continuous sequence in input and also starts from the same position in input i.e. 5, hence it is not clear that whether the user always wants to extract substring started from position 5 in input, or the user have some other desired intent (extract something after space character). In terms of DSL, mostly this kind of ambiguity can be possible because of the operators which allow constant positions to detect a position of substring vs operators which uses regex or split-based operation to extract substring. Formally, we can define this property as, given an \emph{example} (\emph{($I_{1}, O_{1}),(I_{2}, O_{2}), ..., (I_{l}, O_{l})$}), it satisfies this property when the continuous sequence of string in an output matches to the same continuous sequence of string in the corresponding input and it has a start or end always at the same position.

\subsubsection{Exact Match Ambiguity} 
This kind of ambiguity can happen when the output substring of all the I/O pairs can be extracted by applying the same DSL operator on corresponding input annotations and extracted output substring across annotations have the same string value. For example, in Table \ref{tab:Properties_Example}, example 3, following substring or continuous sequence in output ``11" and ``11" are extracted from the similar continuous sequence in input and also have the same string value i.e. 11, hence it is not clear that whether the user always wants to have constant value 11 in output or the user want to extract this value from the input string. In terms of DSL, mostly this kind of ambiguity can be possible because of the operators which allow constant positions to detect a position of substring vs operators like split/substring which allow values to be extracted from the string itself. Formally, we can define this property as, given an \emph{example} (\emph{($I_{1}, O_{1}),(I_{2}, O_{2}), ...., (I_{l}, O_{l})$}) satisfies this property when the continues sequence of string in an output matches to same continuous sequence of string in input and have same value.

\subsubsection{Similar in Token Type Ambiguity} 
This kind of ambiguity can happen when the output substring of all the I/O pairs can be extracted by applying the same DSL operator on corresponding input and extracted output substring across I/O pairs is of the same type. For example, in Table \ref{tab:Properties_Example}, example 4, following substring or continuous sequence in output ``123" and ``53" are extracted from the similar continuous sequence in input and also have the same value type, hence it is not clear that whether the user always wants to extract the same data type value or something else. Mostly, three types of tokens, \emph{Alphabet Tokens} which consists of all uppercase and lowercase English alphabets, \emph{Numeric Tokens} which consists of digits from 0 to 9 and \emph{Special-Character Tokens} which consists of all printable special characters on the keyboard, are possible. Hence, we say that an \emph{example} satisfies a similar token type if all its continuous substring in outputs are either all \emph{Alphabet Tokens} or all \emph{Numeric Tokens} or all \emph{Special-Character Tokens}. In terms of DSL, mostly this kind of ambiguity can be possible because of the operators which allow a specific set of regex positions to detect a position of substring vs operators like split/substring which allows values to be extracted from string itself. Formally, we can define this property as, given an \emph{example} satisfies this property when the continues sequence of string in an output matches to same continuous sequence of string in input and have same value type. 

\subsubsection{Repeating Characters Ambiguity}
This kind of ambiguity can happen when the output substring of all the I/O pairs can be extracted by applying the same DSL operator on corresponding input annotations and have multiple instances of that output substring is possible in input. For example, in Table \ref{tab:Properties_Example}, example 5, following substring or continuous sequence in output ``1" and ``2" can be extracted from two similar positions from the input. Those positions can be defined by any low-level operators like constant positions, regex, or high-level operators like split, etc. In this case, that common substring is possible at two constant positions in input i.e. positions 3 and 9. Hence it is not clear that whether the user wants to extract a substring from position 3 or 9. This is DSL independent ambiguity, which can happen because the user provided the samples in the way that it internally it generating such kind of ambiguity. Formally, we can define this property as, given an \emph{example} satisfies this property when the continues sequence of string in an output matches to multiple instance of continuous sequence of string in input.
\subsection{Problem Formulation}
Given a set of $l$ input-output annotations (\emph{($I_{1}, O_{1}),.., (I_{l}, O_{l})$}), and a set of $p$ properties (\emph{$P_{1}, P_{2}, .., P{p}$}) which can help to detect multi-intents in I/O annotations. The goal of this task is to answer the question ``Is there any multi-intent or ambiguity present in I/O samples", if yes, what kind of ambiguities exist. In this paper, $p$ is set to 5, as we designed and discussed 5 properties in the last section that can hinder the generalization of PBE systems. To learn to detect these sets of ambiguities, we design the multi-tasking attention-based deep neural network model.

We first generate a set of I/O annotation examples corresponding to each of the five ambiguous properties. We refer to a single I/O pair (\emph{$I_{1}, O_{1}$}) as a sample and a group of I/O pairs to learn the program using any PBE system as an \emph{example}. Here, $l$ denotes the total samples (I/O pair) used for each example. In this work, we used $l=3$, which means in each example, we have three I/O samples. One example can have multiple properties issues also. Intuitively, the goal of our proposed task is to detect the ambiguities in the user-provided I/O annotations so that the user resolves these ambiguities by adding the new or modifying the existing samples. This will enable PBE systems to generate a single intent program that performs as desired on the unseen samples.  

In the proposed framework, we train a multi-tasking attention-based deep neural network model as shown in Figure \ref{fig:model} to learn the ambiguities as expressed in the I/O \emph{examples}. We define each task as a formulation to learn one type of ambiguity. Consequently, the proposed framework solves the five tasks at a time corresponding to ambiguity detection for five different properties. Our model follows an encoder-decoder architecture where the encoder is shared among all the tasks and the decoder is independent for each task. We pose this problem as a multi-class classification problem. Each \emph{example} is classified against five ambiguous properties as positive or negative, where positive means that the \emph{example} is ambiguous for that property and negative means that it is not ambiguous.

\vskip 0.2in
\noindent \textbf{Model Architecture} - We model the proposed framework shown in Figure \ref{fig:model} for detecting ambiguities through a hard-parameter sharing paradigm for multi-task learning. As shown in Figure \ref{fig:model}, the proposed framework consists of three modules, Common Encoder, Task-Specific Modules, and the Loss module. We discuss each of these modules in subsequent subsections.

\subsubsection{Common Encoder}
\begin{figure}[t]
  \centering
  \includegraphics[width=\linewidth]{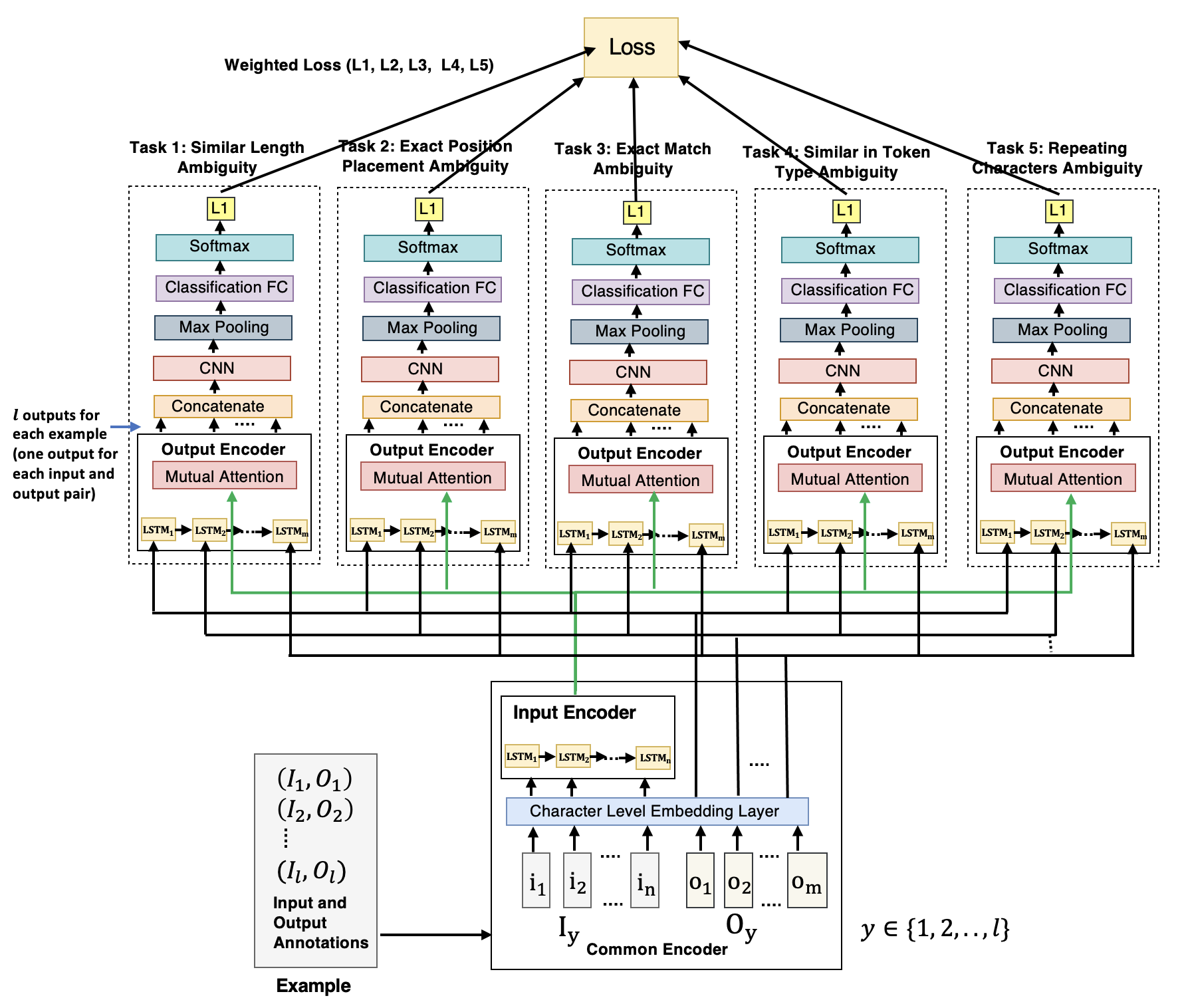}
  \caption{Proposed multi-tasking deep attention based NN architecture to detect multi-intent issues in I/O pairs.}
  \label{fig:model}
  \vspace{-6mm}
\end{figure}

This module is used for encoding the raw I/O strings (see Figure \ref{fig:model}) and consists of two sub-modules:
    \begin{itemize}[leftmargin=*]
        \item \textbf{Character Level Embedding Layer} - This layer maps each character of the I/O pairs in each \emph{example} to a 128-dimensional learning space. Given an input (\emph{$i_{1},.., i_{n}$}) and an output string (\emph{$o_{1},.., o_{m}$}) consisting of a sequence of characters of length $n$ and $m$ respectively, this layer outputs a list of character embedding. Here, n refers to the maximum length of input among all the examples in the dataset. The input strings which are smaller than the maximum length are appended with <pad> tokens to make their length equal to $n$. The <pad> tokens specify that the current character does not signify the original string but marks the end of it or is used to make all sequences of the same length so that the deep learning tensor computations are easier. 
        A similar procedure is followed with the output strings, where the maximum length of output among all the examples in the dataset is $m$. 
        Each character {$i_t$} and {$o_s$} in the input and output sequence is mapped to the 128-dimensional raw embedding $e_{i_{t}}$ and $e_{o_{s}}$ respectively via a randomly initialized and trainable embedding matrix, where $t \in \{1,..n\}$ and $s \in \{1,..m\}$.
        \item \textbf{Input Encoder} - This layer uses LSTM representations \cite{hochreiter1997long} applied on the embedding $e_{i_{t}}$ of the inputs of each \emph{example} as shown in Equation \ref{eq:1}. This layer helps to learn the sequential dependencies of the characters of the inputs. It takes the input embedding of each character  $e_{i_{t}}$ and passes them through a LSTM layer consisting of n separate LSTM cells with a hidden vector size of 512 as shown in Equation \ref{eq:1}.
\begin{equation}\label{eq:1}\footnotesize
\overrightarrow{h_{i_t}} = LSTM(e_{i_{t}},\overrightarrow{h_{i_{t-1}}}), t \in ({1...n})
\end{equation}
\end{itemize}

Hence, the Common Encoder takes I/O pair as input and produces two output representations, the raw 128-dimensional embedding of each character in the output sample and the LSTM encoded embedding of the Input sample in the I/O pair. These embeddings will be generated for each I/O pairs in an example. The outputs of the Common Encoder are then utilized by next Modules.

\subsubsection{Task Specific Modules}
These modules are designed for the detection of each ambiguity property. We have 5 such modules (one for each ambiguity) with a similar structure, which process the inputs obtained from the common encoder. Each Task-Specific Module contains an Additive Attention Output Encoder, Concatenation Layer, Convolution Neural Networks \& Pooling Layer, and Softmax Layer as classification layer. The weights across all these 5 task-specific modules are not shared with each other.
\begin{itemize}[leftmargin=*]
\item \textbf{Attention Output Encoder} - 
In our architecture, we use additive attention mechanism \cite{bahdanau2014neural} to selectively impart more importance to the part of the input which has more influence on the output characters and hence obtain better output sample encoding. Specifically, this layer computes the additive attention $a_{e_{o_{s}}}$ of a single embedded output character $e_{o_{s}}$ with respect to the encoding of all the input characters  {$h_{i_{1..n}}$} as shown in the equations \ref{eq:2} and \ref{eq:3}. For this, we pass the output from the Input Encoder to the Attention Output Encoder which first computes the attention weights $\alpha_{s_{1..n}}$ as shown in eq. \ref{eq:2} and the corresponding attention vector $a_{e_{o_s}}$ as shown in eq. \ref{eq:3} for each output character $O_s$ with respect to all input characters in the I/O pair. Here, $W_a$ and $U_a$ are the learnable weight matrices. $W_a$ corresponds to the output embeddings vector $e_{o_s}$ and $U_a$ corresponds to the input encodings matrix $h_{i_{1..n}}$. $V_a$ is the learnable vector.
The attention output $a_{e_{o_{s}}}$ is concatenated with the output embedding $e_{o_s}$ to give $c_{o_s}$ as shown in equation \ref{eq:4} and is passed through an LSTM layer with hidden vector size 512 as shown in eq. \ref{eq:5}.  
\begin{equation}\label{eq:2}\footnotesize
    \alpha_{s_{1..n}} = {V_a}tanh({W_a}{e_{o_s}} + {U_a}h_{i_{1..n}}), s \in ({1..m})
\end{equation}
\begin{equation}\label{eq:3}\footnotesize
   a_{e_{o_{s}}} = \Sigma_{t} \alpha_{s_{t}} h_{i_{t}}, t \in ({1..n}), s \in ({1..m})
\end{equation}
\begin{equation}\label{eq:4}\footnotesize
{c_{o_s}}= [ a_{e_{o_s}},e_{o_s}], s \in ({1..m})
\end{equation}
\begin{equation}\label{eq:5}\footnotesize
\overrightarrow{h_{o_s}} = LSTM(c_{o_s},\overrightarrow{h_{o_{s-1}}}), s \in ({1..m})
\end{equation}
The Attention Output Encoder outputs $m$ different LSTM encodings $h_{O_{1..m}}$ for each output string of length $m$ in $l$ I/O pairs, which further passed to the next Layer.

\item \textbf{Concatenation Layer} - 
For this, we concatenate the $l$ encodings corresponding to $l$ I/O pairs for each example. Detecting ambiguity is possible only by analyzing all the I/O pairs in a given \emph{example} and not just one I/O pair. These encodings are obtained from the Attention Output Encoder in a row-wise manner as shown in equation \ref{eq:6}. Here, $h_{1_{o_s}}$ refers to the attention-encoded output of the $s^{th}$ character of the Output $O_1$ from the first I/O pair. Similarly, $h_{l_{o_s}}$ refers to the attention-encoded output of the $s^{th}$ character of the Output $O_l$ from the $l^{th}$ I/O pair.
\begin{equation}\label{eq:6}\footnotesize
   q_s = concat(h_{1_{o_s}}, h_{2_{o_{s}}}, ..., h_{l_{o_{s}}}), s \in ({1..m})
\end{equation}
\begin{equation}\label{eq:7}\footnotesize
   Q = [q_1, q_2, ...., q_{m-1}, q_{m}]
\end{equation}
The output of the Concatenation Layer is a matrix $Q$ as shown in eq. \ref{eq:7}. There are a total of $m$ different rows in the matrix corresponding to the $m$ characters of the Outputs in an I/O pair. More specifically, each row of the matrix represents the character-level concatenation of the output encodings from $l$ different examples. This matrix is then passed into the next layer.

\item \textbf{Convolution Neural Network and Pooling Layers} - 
Convolution Neural Networks (CNNs) are used for finding local dependencies in features. In our architecture, CNNs help us to capture the dependencies between adjacent characters and subsequent encoded Outputs of the I/O pairs. The input to the CNN layer is the matrix $Q$ for each \emph{example} that we obtain from the Concatenation Layer. In this layer, we apply 2-dimensional convolution operations with 512 output channels where each channel contains a kernel of dimension (2, $l$*512) on the input from the concatenation layer. We then applying MaxPooling on the outputs of the CNNs across each channel to obtain a single vector $r$ of size 512 dimensions for the I/O pairs in an \emph{example} as seen in eq. \ref{eq:8}. This 512 size vector is then passed into the next Layer.
\begin{equation}\label{eq:8}\footnotesize
   r = MaxPool2D(Conv2D(Q))
\end{equation}

\item \textbf{Classification Layer} - Classification Layer is a fully-connected dense layer with 2 neurons corresponding to either the positive or negative class for each ambiguous property classification to give the classification logits $u$. This is shown in equation \ref{eq:9} where {$W_{f}$} and {$b_{f}$} are the weight matrix and the bias vector respectively. 
\begin{equation}\label{eq:9}\footnotesize
       u = W_{f} r + b_{f}
\end{equation}
Classification logits from the Classification Layer are then passed through the Softmax Layer.

\item \textbf{Softmax Layer} - This layer applies the softmax activation function on the classification logits to obtain a probability distribution $p$ over the prediction classes (ambiguous properties) as shown in equation \ref{eq:10}. Here, $z$ is used for indexing a single class among the positive and the negative classes. 
\begin{equation}\label{eq:10}\footnotesize
    p = \frac{exp(u_{z})}{\Sigma_z exp(u_{z})}, z \in ({0, 1})
\end{equation}
\end{itemize}
\subsubsection{Loss Calculation}
The proposed multitask learning framework uses Cross-Entropy loss between the original and predicted labels as the objective function for all the five task-specific modules. Equation \ref{eq:11} denotes the loss from the $k^{th}$ task-specific module. We use $k$ to index the task-specific modules. $p_k$ is the predicted probability distribution for the $k^{th}$ task-specific module. $y_k$ is the original probability distribution for the $k^{th}$ task-specific module.
We obtain the final loss $L$ by taking a weighted sum of the individual losses $L_k$ of each of the task-specific modules as shown in equation \ref{eq:12}. Here $w_k$ is the weight corresponding to the $k$th Loss $L_k$.  
    \begin{equation}\label{eq:11}\footnotesize
 L_k = -\Sigma[{y_k}log{p_k} + (1-{y_k})log(1-{p_k})]
\end{equation}
\begin{equation}\label{eq:12}\footnotesize
 L = w_1*L_1 + w_2*L_2 + w_3*L_3 + w_4*L_4 + w_5*L_5
\end{equation}

\section{Results and Discussions}
\subsection{Dataset Creation} 
We created a dataset corresponding to the five different ambiguous properties discussed in Section \ref{prop}. We have written different regexes satisfying each ambiguous property based on a fixed Domain Specific Language (DSL). For each ambiguity property, the regexes generate several examples, and each \emph{example} consists of 3 I/O pairs. We consider uppercase English characters, lowercase English characters, digits from 0 to 9, and all printable special characters. We generate a total of 100002 individual samples, grouped in an \emph{example} of 3 samples, to finally produce 33334 \emph{examples} per ambiguous property. In the next few subsections, we describe the procedure of generating the dataset for each ambiguous property. Table \ref{tab:Properties_Example} shows examples corresponding to each property.

\subsubsection{Similar Length Ambiguity} For each output substring in an \emph{example}, we chose a length from a range of 2-9 characters. We limit the output substrings to a maximum of 4 for each sample. Each output substring will contain a mixture of lowercase, uppercase English alphabets, and digits from 0-9. We add random strings on the front and the back of each output substring to construct the input string. Similarly, we do this for other output substrings, and finally, combine the I/O substrings to make it a single I/O pair. We repeat the above process by fixing the output substrings size across the samples in a single example and combine those I/O pairs to make a single \emph{example}. In our case, we use a set of three I/O pairs in a single \emph{example}. 
We illustrate the process of creating I/O pairs through the following example. In the first step, we first assume an output substring of length three for sample-1 is ``abc", for sample-2 is ``klp" and for sample-3 is ``12j". In the second step, we add random I/O strings before and after the first output substring for sample-1 ``dfg1\#\textbf{abc}\#2311", sample-2 ``era\#\textbf{klp}\#hj1", and sample-3 ``h2ral\#\textbf{12j}\#klj23jk". In the third step, we create a new output substring that can follow this similar length property or not. We then repeat the second step, for example, let us assume that the second output substring is of varied length, let's say ``hjuk", ``puefhkj", and ``jf16hsk". Now, either we append this directly to the input with some delimiter or first add some other random string before or after this string. In this case, we append this directly using delimiter ``@", so final input strings become ``dfg1\#abc\#2311@hjuk", ``era\#klp\#hj1@puefhkj" and ``h2ral\#12j\#klj23jk@jf16hsk". We can combine the output substrings using any character or directly. In this example, we are combining it directly which leads to the following output samples corresponding to the input samples - `abchjuk", ``klppuefhkj" and ``12jjf16hsk". We can repeat the same process by generating more output substrings for an \emph{example}.

\subsubsection{Exact Position Placement Ambiguity} The process of \emph{example} generation for this ambiguity will remain almost the same as the ``Similar Length Ambiguity" property. The only change is that instead of fixing output substring length across samples, we will fix the output substring's position in the input string.

\subsubsection{Exact Match Ambiguity} - In this case, the process differs with respect to output substring value. The output substring value across the I/O pairs within the same \emph{example} will remain the same. This property inherently also satisfies the Similar Length Ambiguity.

\subsubsection{Similar in Token Type Ambiguity} In this case, the process differs with respect to output substring type. That is, the output substring's token-type across the I/O pairs within same \emph{example} will remain the same. In our work, we define two types of token-types viz. alphabets and numerals. More specifically, the two categories of similar token types are when, either the output strings contain only the uppercase and lowercase alphabets or only digits from 0-9.

\subsubsection{Repeating Characters Ambiguity} In this case, the output substring exists (or repeats itself) at multiple positions in the input.

\subsection{Ablation Studies}
We compare the results of two major variations in the proposed framework : (a) two different loss functions - Cross-Entropy and Focal Loss, and (b) the importance of each layer by removing it from the framework. We consider the model in Figure \ref{fig:model} as the main model. This model is referred to as \textbf{Our} in the results table. We carry out various ablation studies of the proposed model by removing various components to ascertain the role played by each component in the model. These models are are discussed below. 

\subsubsection{Our\_No\_CNN:} In this setup, we remove the CNN and the MaxPool layers from the proposed model architecture and only pass the concatenated output encodings to the classification layer. 

\subsubsection{Our\_No\_AM} In this setup, we remove the Attention Mechanism from the proposed model. We retain the same output encoder but set the attention weights for each output characters over all input characters is equal to 1 while calculating the attention vector. 

\subsubsection{Our\_GRU:} In this, we replace all LSTM layers and cells with GRU \cite{cho2014learning} cells in the proposed architecture. We retain the same overall architecture and keep the GRU hidden size equal to 512.

\begin{table*}[t]\scriptsize
    \begin{minipage}{.99\linewidth}
      \centering
      
      \begin{tabular}{ |m{1.2cm}|l|l|l|l|l|l|l|l|l|l|l|l|l|l|l|l| }
      \hline
           \textbf{Loss} & \textbf{Approach} & \multicolumn{3}{c|}{\textbf{Similar Length}} & \multicolumn{3}{c|}{\textbf{Repeating Characters}} & \multicolumn{3}{c|}{\textbf{Exact Match}} & \multicolumn{3}{c|}{\textbf{Similar in Token Type}}& \multicolumn{3}{c|}{\textbf{Exact Position Placement}}  
            \\ \hline
 & & NF1  & PF1 & Acc& NF1  & PF1 & Acc& NF1  & PF1 & Acc& NF1  & PF1 & Acc& NF1  & PF1 & Acc\\\hline
            Focal Loss         & Our\_No\_CNN       & 0.91      & 0.85      & 0.88  & 1      & 0.98      & 0.99 & 0.50     & 0.42      & 0.46  & 0.89      & 0.08      & 0.81   & 0.56      & 0.45      & 0.51 \\\cline{2-17} 
            & Our\_No\_AM & 1         & 1         & 1& 1        & 1        & 1 & 1         & 1        & 1 & 0.89         & 0.08         & 0.81 & 0.87         & 0.69         & 0.82\\\cline{2-17} 
            & Our\_GRU         & 0.54      & 0.68      & 0.62   & 1         & 1        & 1    &1      & 1     & 1 & 0.89      & 0.00     & 0.80  & 0.87      & 0.69     & 0.82 \\\cline{2-17} 
             & \textbf{Our}         & 1         & 1         & 1 & 1         & 1         & 1 & 1         & 1         & 1   & 0.89         & 0.08        & 0.81& 0.87         & 0.69        & 0.82  \\\cline{1-17} 
             \multirow{3}{1.2cm}{\textbf{Cross Entropy Loss}} & Our\_No\_CNN       &      0.56     &      0.69     &  0.6  &      1     &      0.98     &  0.99&      0.56     &      0.45     &  0.51 &      0.92     &      0.42     &  0.85 &      0.95     &      0.77     &  0.91\\\cline{2-17} 
              & Our\_No\_AM & 0.89         & 0.88         & 0.80 & 1         & 1         & 1 & 1         & 1         & 1 & 0.90         & 0.13         & 0.81 & 0.93        & 0.73         & 0.89  \\\cline{2-17} 
              & Our\_GRU          & 1         & 1         & 1 & 0.89         & 0.88        & 0.80 & 0.98         & 0.93        & 0.97 & 0.89         & 0.08        & 0.81  & 0.76         & 0.44        & 0.66\\\cline{2-17} 
               & \textbf{Our }        &      1     &      1     &  1  &      1     &      0.98     &  0.99 &      1     &      1     &  1 &      1     &      1     &  1 &      1     &      1     &  1\\   \cline{1-17} 
        \end{tabular}
    \end{minipage}
    \caption{Tables show the results of the proposed framework for each type of ambiguity. NF1, PF1, Acc denotes the F1 score for samples labeled false, labeled as true and accuracy respectively of the proposed framework for that particular property.} 
     \label{quantR}
\vspace{-5.5mm}
\end{table*}

\subsection{Discussions}
\subsubsection{Quantitative Results} 
In Table \ref{quantR}, we compare the results of the proposed framework with two different loss functions, Cross-Entropy, and Focal Loss. Also, we provide a quantitative analysis highlighting the importance of each layer. For this, we first remove the layer from the task-specific modules and then report the performance of the same. We show the property-wise performance in Table \ref{quantR}. From the result table, we can see that overall Cross-Entropy is performing better than the Focal Loss. The model has trained with 26,667 \emph{examples} corresponding to each ambiguous property for 100 epochs with a batch size of 5 per epoch. We set the weights of each loss corresponding to five different ambiguity tasks equally to 1. We report the results on the 6,667 \emph{examples} test set. 

The main model, denoted by \textbf{our}, performs better than the other variations of the proposed framework when using the same loss metric. We can also observe that by removing the attention layer from the main model, the performance of the model got decreased by 10-20\% for most of the cases, which highlights the need for an attention layer. A similar kind of pattern can be observed when we remove the CNN layer from the main model. In some cases, performance got dropped to around 50\%. Also, we can observe that removing the CNN layer makes the model more worse as compared to removing the attention layer. This shows that the CNN part of the architecture plays an important role in ambiguity detection. Also, we can see a  significant drop in performance in most of the cases, if we replace the LSTM units with GRU units. The reason for the same is that LSTM units are able to capture context better than GRU because a sufficient number of samples are provided to our model to learn the context. When a sufficient number of samples are available for training, we can expect the LSTM model to learn the context better than GRU \cite{gruber2020gru}. Hence, this analysis shows the importance of different layers in our proposed framework.

Combining all these layers, makes the system perform almost 100 percent accurately on the test set, which shows that these ambiguities can be easily learned if we define the architecture which can capture context, interrelationships, and attention of output on input. In some cases, we observe that other variations are also giving perfect results which highlight that for those properties simpler network can also be generalizable on unseen test data.

\begin{figure}[t]
  \centering
  \includegraphics[scale=0.10]{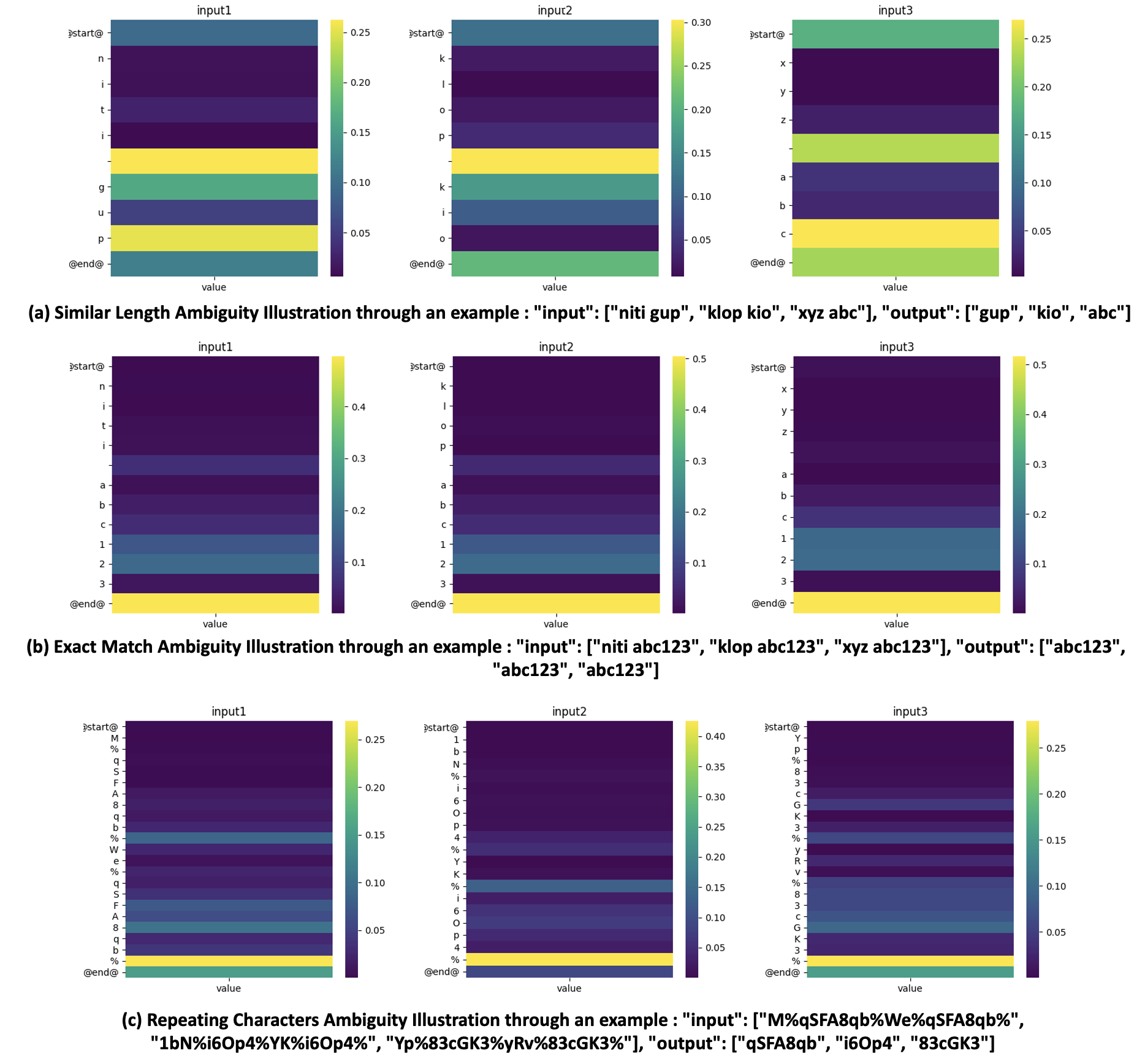}
  \caption{Illustration of ambiguity properties through Saliency Maps. The lighter the color higher its `importance".}
  \label{fig:saliency}
  \vspace{-6.5mm}
\end{figure}

\begin{table*}[t] \scriptsize
\centering
\begin{tabularx}{\linewidth}{|l|m{1.5cm}|l|m{3.0cm}|L|L|L|L|L|L|L|L|}
\hline
     \textbf{S.No.}&\multicolumn{3}{|c}{\textbf{Train}}& \multicolumn{4}{|c}{\textbf{Test without fourth sample}}& \multicolumn{4}{|c|}{\textbf{Test with fourth sample}}\\\cline{2-12}\hline
    &\textbf{Input }& \textbf{Output}  &\textbf{Ambiguities Detected }& \textbf{Input}& \textbf{PROSE Output}&\textbf{Excel Output}&\textbf{GT Output}& \textbf{Input}& \textbf{PROSE Output}&\textbf{Excel Output}&\textbf{GT Output} \\\hline
    &ABCD\_12 &12 & \multirow{3}{2.8cm}{(a) Similar in Token Type} & & && & & &&\\
    1&BDJ\_535 &535 && B\_DS2345 & 2345 & DS2345 & DS2345 & B\_DS2345 & DS2345 & DS2345 & DS2345 \\
    &GE\_443 &443& & & && & & &&\\
    &\textbf{AK\_B121} &\textbf{B121}& & & && & & &&\\
    \hline
    &ABC232 &232& \multirow{2}{2.8cm}{(a) Similar Length \\ (b) Similar in Token Type} & & && & & &&\\
    2&BD345 & 345 &  & ABC23B & 23 & 23B &23B& ABC23B & 23B & 23B &23B\\
    &GSS564 & 564 & & & && & & &&\\
    &\textbf{ABC12A }& \textbf{12A} && & && & & &&\\
    \hline
    &baiTree &bai& \multirow{3}{2.8cm}{(a) Similar Length \\ (b) Similar in Token Type \\ (c) Exact Position Placement} & & && & & &&\\
    3&paiRcr & pai &  & lmptFree & lmpt & lmpt &lmp& lmptFree & lmp & lmp &lmp\\
    &cmiPrent & cmi & & & && & & &&\\
     &\textbf{cm12Prent }& \textbf{cm1 } && & && & & &&\\
    \hline
    &sf334.12 & 334 & \multirow{2}{2.8cm}{(a) Similar in Token Type \\ (b) Exact Position Placement}  & & && & & &&\\
    4&fg42.456 &42 & & fsd456.63 & 456 & 456 &d456 & fsd456.63 & d456 & d456 &d456 \\
    &gd12.44 & 12 &  & & && & & &&\\ 
     &\textbf{khb79.44} & \textbf{b79}   && & && & & &&\\
    \hline
    &Mohan Mrinesh & Mr. Mohan & \multirow{4}{2.8cm}{(a) Similar Token Type\\(b) Similar in Exact Match\\(c) Similar Length \\ (d) Exact Position Placement} &Ramesh &Ba.& Ba. &Mr.  &Ramesh  & &&\\
    5&Abhil Mritha & Mr. Abhil&  &  Bansal &  Ramesh & Ramesh &Ramesh &  Bansal & Ramesh & Ramesh &Ramesh\\
    &Johny Mrendash & Mr. Johny &  & & && & & &&\\ 
    &\textbf{John Sharma} & \textbf{Mr. John} &  & & && & & &&\\
    \hline
    &07-07-1999 & 07-99 & \multirow{4}{2.8cm}{(a) Similar in Token Type\\ (b) Similar Length \\(c) Repeating Characters \\(d) Exact Position Placement} & & && & & && \\
    6&02-02-1955 & 02-55 &  & 09-07-1995 & 09-95 & 09-95 &07-95& 09-07-1995 & 07-95 & 07-95 &07-95 \\
    &10-10-2002 & 10-02 & & & && & & && \\
    &\textbf{10-11-2002} & \textbf{11-02} &  & & && & & &&\\
    
    \hline

    \end{tabularx}
    \caption{Illustration of detected ambiguities by the proposed framework. The table also highlights that if we generate the program through existing widely used PBE systems like PROSE and Excel without fixing the detected ambiguities then it can lead to wrong output on test samples (Test without fourth example column). We also showed change in output (Test with fourth example column) of PROSE and Excel systems after modifying I/O samples (highlighted in bold).}
        \label{tb:casestudy}   
    \vspace{-6.5mm}
\end{table*}

\subsubsection{Saliency Maps}
For better understanding the predictions of the proposed model, we used the integrated gradients \cite{sundararajan2017axiomatic} based saliency on the inputs of the \emph{examples} for visualization. We use three properties (similar length, exact match, and repeating characters) to illustrate the predictions of the learned model as shown in Figure \ref{fig:saliency}. For each of these properties, we use one example (three I/O samples) to visualize the saliency maps. Also, we use a single substring in output just for ease of visualization, as this visualization becomes more complex to interpret if we have multiple substrings in output.

The first row in the Figure \ref{fig:saliency} denotes the saliency maps corresponding to the Similar Length Ambiguity property for the I/O pair - \{"input": ["niti gup", "klop kio", "xyz abc"], "output": ["gup", "kio", "abc"]\}. From Figure \ref{fig:saliency} (a), we can see that in all the inputs, more importance (shown by lighter colors with high values) is given to the characters which mark the beginning and the end of the part of the string (``gup", ``kio", and ``abc") which belongs to the output. That is, we can see that a higher saliency score is associated with the hyphen and the @end symbol which mark the beginning and the ending of the output string. Hence, we can conclude that the model is able to learn the Similar Length Ambiguity property.

The second row illustrates the saliency maps for the Exact Match Ambiguity property for the I/O pair - \{"input": ["niti abc123", "klop abc123", "xyz abc123"], "output": ["abc123", "abc123", "abc123"]\}. Here, it can be seen that, on average, more importance is given to the part of the input which contains the output as compared to the one which does not. That is, the characters corresponding to abc123 have higher saliency values as compared to the other parts like niti, klop, and xyz in the three inputs respectively. Hence, we can conclude that the model is able to recognize the output strings clearly and hence correctly classifying them. 

The third row shows the saliency maps for the Repeating Characters Ambiguity for the I/O pair - \{"input": ["M\%qSFA8qb\%We \%qSFA8qb\%", "1bN\%i6Op4\%YK\%i6Op4\%", "Yp\%83cGK3\%yRv\%83cGK3\%"], "output": ["qSFA8qb", "i6Op4", "83cGK3"]\}. It can be noticed that the characters in the output string have higher saliency values on an average in the input in their second repetition as compared to their first occurrence. This shows that the model is able to well recognize the repeated characters and hence correctly classify them. We have observed similar kinds of patterns for the other ambiguities.

\subsubsection{Case Study: Impact of detecting multiple intents and correcting them before building PBE systems}
In this section, we discuss that how the presence of ambiguity in input and output annotations can affect the output of widely used tools like PROSE \cite{polozov2015flashmeta} and Microsft Excel. Table \ref{tb:casestudy}, shows the different ambiguities detected by the proposed system on 6 examples, and also shows that whether existing PBE systems will able to learn correct intent or not using those sets of I/O pairs. For each example, the user provides three I/O samples to convey the desired intent. However, as we can see from the ambiguities detected column that each of these examples has some kind of ambiguities or multi-intent issues. Effect of the same can be reflected in a mismatch of PROSE/Excel output columns and GT output column. This shows the need for the framework which helps to figure out the multi-intent quality issues in annotation before generating program through any PBE systems.

In the first example in Table \ref{tb:casestudy}, the system detects ``Similar in Token Type Ambiguity", because substring (only one substring  exist in this case) across the outputs have the same token type. This can lead to multiple intent issues of whether the user wants to extract everything after ``\_" irrespective of the token/data type, or user is just interested in specific numeric data type content for this case. Same multi-intent confusion can be reflected in the output of two different PBE systems on an input ``B\_DS2345" - (a) PROSE output is ``2345", that means the PROSE framework learn to extract numeric content after ``\_", and (b) Excel output is ``DS2345" which means that excel learns to extract all the content after ``\_". So, it is good if the user can first analyze the detected ambiguity and if that ambiguity holds for a user's actual intent, then the user can accordingly either provide new samples or change the existing samples. Like for the first example, the user intent is to extract everything after ``\_" and also detected ambiguity is of similar token type. So, the user can now either modify or add one new sample where the extracted output string also has non-numerical characters. With this new additional I/O sample (highlighted in bold) provided by a user, after analyzing the detected ambiguity, both PROSE and EXCEL are able to learn correct intent. This is reflected through the output columns i.e. the value of these columns is the same as the GT column (see Table \ref{tb:casestudy}) .

Similarly, if we analyze the fifth example in Table \ref{tb:casestudy}, the system detects multiple ambiguities. Exact Position, Similar Length, and Similar in Token Type ambiguities exist for both the output substrings (Mohan/Abhil/Johny and Mr.). Similar in Exact Match Ambiguity exists only for ``MR" substring in the output. For the first output substring (Mohan/Abhil/Johny), the user is fine with Exact Position and Similar in Token Type Ambiguity. However, the user wants to add a new example to remove the Similar Length Ambiguity. Similarly, for the second output substring, the user is fine with all the detected ambiguities except Exact Position Placement Ambiguity, because the user's goal is not to extract this information from the input string, the user wants to add that as a constant string in the output. So, after analyzing these properties, the user can provide new samples which will remove these ambiguities to learn the correct intent. Also, we can see from the table that due to these ambiguities both PROSE and EXCEL system learn the intent wrongly. However, after analyzing the ambiguities, the user provided the new sample as shown in Table \ref{tb:casestudy}. This new sample helps the system to learn the correct intent, which can be seen through the correct output on the test data.

Similarly, by providing new sample as shown in Table \ref{tb:casestudy} for other examples, the user will be able to resolve the multi-intent quality issue and also be able to learn the correct intent through existing PBE frameworks. This shows the effectiveness of our proposed framework to detect ambiguity in PBE systems specifically in the string transformation domain.

\section{Related Work}
Task-specific string transformation can be achieved via both program synthesis and induction models. Induction-based approaches obviate the need for a DSL since they are trained to generate required output directly from the input string and used in tasks like array sorting \cite{mnih2014recurrent}, long binary multiplication \cite{kaiser2015neural}, etc. However, induction models are not feasible for the string transformation domain as they require to be re-trained for each task and have lower generalization accuracy on unseen samples than synthesis models \cite{devlin2017robustfill}. In literature, both neural-guided-based and symbolic-based approaches have been widely used for program synthesis.

Several neural-guided approaches have been proposed in the last few years for program synthesis \cite{devlin2017robustfill, shu2017neural, parisotto2016neuro}. A sequential encoder-decoder network to infer transformation programs that are robust to noise present in input-output strings, where the hand-engineered symbolic systems fail terribly is proposed in \cite{devlin2017robustfill}. A different variant of an encoder-decoder network where input-output string encoders are not cascaded but work in parallel to infer program sequences is proposed in \cite{shu2017neural}. In \cite{parisotto2016neuro}, a novel neural architecture consisting of a R3NN module that synthesizes a program by incrementally expanding partial programs is used. These networks can be trained end-to-end and do not require any deductive algorithm for searching the hypotheses space. However, they do not guarantee that inferred programs are consistent with the observed set of input-output pairs and also, training on synthetically generated datasets results in poor generalizability on real-world tasks.

Symbolic Program Synthesis approaches operate by dividing required transformation tasks into sub-tasks and searching the hypothesis space for regex-based string expressions to solve each of them. However, smart search and ranking
strategies to efficiently navigate the huge hypothesis search space require significant engineering effort and domain knowledge. One of the earliest attempts to solve the problem
of program synthesis pioneered the Flash-Fill algorithm designed to infer specification satisfying string transformation program in the
form of Abstract Syntax Trees (AST) \cite{gulwani2011automating}. The
PROSE system from \cite{polozov2015flashmeta} employs several hand-crafted heuristics to design ranking functions for deductive searching. Systems like PROSE though perform well on tasks similar to the previously encountered tasks but face a generalizability issue when exposed to new unseen tasks. This is also demonstrated in Table \ref{tb:casestudy} where the system infers one intent which is satisfied in the seen examples but fails on new unseen test data. Since PBE systems for string transformations rely on input and output annotations, it is necessary to provide non-ambiguous input and output samples to them. There is no work existing in the literature that talks about finding the ambiguities or multiple intent-based quality issues in input and output annotations, and providing that information to the user so that the user can look for those detected ambiguities and accordingly modify existing samples or provide new samples. This kind of system will help to capture the user's intent more clearly and make the system automatically generalizable on unseen data. Hence, in this paper we focused on finding the quality issues in input-output annotations with respect to multi-intent, to learn correct intent.

\section{Conclusion}
This paper aims to solve the problem of detecting ambiguity in the user-provided I/O annotations for PBE systems which leads to the generation of wrong intent programs. To the best of our knowledge, our proposed framework is the first to solve this issue at the input and output annotation level. To solve this, we propose extensible multi-tasking attention-based DNN to find the multiple intents in the I/O samples. We also define a set of generic properties that help in detecting the multiple intents in the annotations. We have done a quantitative analysis of different variations of the proposed model architecture to show the impact of the proposed systems' modules. We have also illustrated the effectiveness of the proposed model through saliency maps and by using an existing PBE system outputs. A natural extension of our work is to use the detected ambiguity properties to automatically generate new input and output samples and to improve the program search space.

\bibliographystyle{siam}
\bibliography{sample-base}

\begin{thebibliography}{10}

\bibitem{alur2013syntax}
{\sc R.~Alur, R.~Bodik, G.~Juniwal, M.~M. Martin, M.~Raghothaman, S.~A. Seshia,
  R.~Singh, A.~Solar-Lezama, E.~Torlak, and A.~Udupa}, {\em Syntax-guided
  synthesis}, IEEE, 2013.

\bibitem{bahdanau2014neural}
{\sc D.~Bahdanau, K.~Cho, and Y.~Bengio}, {\em Neural machine translation by
  jointly learning to align and translate}, arXiv preprint arXiv:1409.0473,
  (2014).

\bibitem{cho2014learning}
{\sc K.~Cho, B.~Van~Merri{\"e}nboer, C.~Gulcehre, D.~Bahdanau, F.~Bougares,
  H.~Schwenk, and Y.~Bengio}, {\em Learning phrase representations using rnn
  encoder-decoder for statistical machine translation}, arXiv preprint
  arXiv:1406.1078,  (2014).

\bibitem{devlin2017robustfill}
{\sc J.~Devlin, J.~Uesato, S.~Bhupatiraju, R.~Singh, A.-r. Mohamed, and
  P.~Kohli}, {\em Robustfill: Neural program learning under noisy i/o}, in
  International conference on machine learning, PMLR, 2017, pp.~990--998.

\bibitem{gruber2020gru}
{\sc N.~Gruber and A.~Jockisch}, {\em Are gru cells more specific and lstm
  cells more sensitive in motive classification of text?}, Frontiers in
  artificial intelligence, 3 (2020), p.~40.

\bibitem{gulwani2011automating}
{\sc S.~Gulwani}, {\em Automating string processing in spreadsheets using
  input-output examples}, ACM Sigplan Notices, 46 (2011), pp.~317--330.

\bibitem{hochreiter1997long}
{\sc S.~Hochreiter and J.~Schmidhuber}, {\em Long short-term memory}, Neural
  computation, 9 (1997), pp.~1735--1780.

\bibitem{kaiser2015neural}
{\sc {\L}.~Kaiser and I.~Sutskever}, {\em Neural gpus learn algorithms}, arXiv
  preprint arXiv:1511.08228,  (2015).

\bibitem{manna1975knowledge}
{\sc Z.~Manna and R.~Waldinger}, {\em Knowledge and reasoning in program
  synthesis}, Artificial intelligence, 6 (1975), pp.~175--208.

\bibitem{menon2013machine}
{\sc A.~Menon, O.~Tamuz, S.~Gulwani, B.~Lampson, and A.~Kalai}, {\em A machine
  learning framework for programming by example}, in International Conference
  on Machine Learning, PMLR, 2013, pp.~187--195.

\bibitem{mnih2014recurrent}
{\sc V.~Mnih, N.~Heess, A.~Graves, et~al.}, {\em Recurrent models of visual
  attention}, in Advances in neural information processing systems, 2014,
  pp.~2204--2212.

\bibitem{parisotto2016neuro}
{\sc E.~Parisotto, A.-r. Mohamed, R.~Singh, L.~Li, D.~Zhou, and P.~Kohli}, {\em
  Neuro-symbolic program synthesis}, arXiv preprint arXiv:1611.01855,  (2016).

\bibitem{polozov2015flashmeta}
{\sc O.~Polozov and S.~Gulwani}, {\em Flashmeta: A framework for inductive
  program synthesis}, in Proceedings of the 2015 ACM SIGPLAN International
  Conference on Object-Oriented Programming, Systems, Languages, and
  Applications, 2015, pp.~107--126.

\bibitem{shu2017neural}
{\sc C.~Shu and H.~Zhang}, {\em Neural programming by example}, in Thirty-First
  AAAI Conference on Artificial Intelligence, 2017.

\bibitem{sundararajan2017axiomatic}
{\sc M.~Sundararajan, A.~Taly, and Q.~Yan}, {\em Axiomatic attribution for deep
  networks}, in International Conference on Machine Learning, PMLR, 2017,
  pp.~3319--3328.

\end{thebibliography}
\end{document}